%% file: paper.tex
\documentclass[10pt,twocolumn,letterpaper]{article}
\usepackage[utf8]{inputenc} 
\usepackage[T1]{fontenc}    
\usepackage{csquotes}

\usepackage{comment}
\usepackage{placeins}
\usepackage{float}

\usepackage[final]{cvpr} 
\input{preamble}

\definecolor{cvprblue}{rgb}{0.21,0.49,0.74}
\usepackage[pagebackref,breaklinks,colorlinks,allcolors=cvprblue]{hyperref}

\begin{document}

\title{Learning Visual Composition through Improved Semantic Guidance}

\author{
    \textbf{Austin Stone}\thanks{Equal contribution.\protect\\ Correspondence: \texttt{\{austinstone,soltau\}@google.com}} \qquad 
    \textbf{Hagen Soltau}\footnotemark[1] \qquad 
    \textbf{Robert Geirhos} \qquad
    \textbf{Xi Yi} \\ 
    \textbf{Ye Xia} \qquad
    \textbf{Bingyi Cao} \qquad
    \textbf{Kaifeng Chen} \qquad
    \textbf{Abhijit Ogale} \qquad
    \textbf{Jonathon Shlens} \\ \\
    {\large Google DeepMind} 
}

\renewcommand{\thefootnote}{\fnsymbol{footnote}}
\maketitle

\renewcommand{\thefootnote}{\arabic{footnote}} 

\input{sections/abstract}
\input{sections/intro}
\input{sections/methods}
\input{sections/results}
\input{sections/related}
\input{sections/discussion}

{
    \small
    \bibliographystyle{ieeenat_fullname}
    \bibliography{paper}
}

\input{sections/appendix}

\end{document}

%% file: preamble.tex
\usepackage{epsfig}
\usepackage{graphicx}
\usepackage{amsmath}
\usepackage{amssymb}
\usepackage{textcomp}
\usepackage{booktabs} 
\usepackage{graphicx}
\usepackage[utf8]{inputenc}
\usepackage{color}
\usepackage{multirow}
\usepackage{lipsum}
\usepackage{enumitem}
\usepackage{tabularx}

\definecolor{baselinecolor}{gray}{.9}

%% file: sections/abstract.tex
\begin{abstract}
Visual imagery does not consist of solitary objects, but instead reflects the composition of a multitude of fluid concepts.
While there have been great advances in visual representation learning, such advances have focused on building better representations for a small number of discrete objects bereft of an understanding of how these objects are interacting.
One can observe this limitation in representations learned through captions or contrastive learning -- where the learned model treats an image essentially as a bag of words.
Several works have attempted to address this limitation through the development of bespoke architectures.
In this work, we focus on simple and scalable approaches. 
In particular, we demonstrate that by improving weakly labeled data, i.e. captions, we can vastly improve the performance of standard contrastive learning approaches. 
Previous CLIP models achieved near chance rate on challenging tasks probing compositional learning. However, our simple approach boosts performance of CLIP substantially and achieves state of the art results on compositional benchmarks such as ARO and SugarCrepe.
Furthermore, we showcase our results on a relatively new captioning benchmark derived from DOCCI. We demonstrate through a series of ablations that a standard CLIP model trained with enhanced data may demonstrate impressive performance on image retrieval tasks.
\end{abstract}

%% file: sections/intro.tex
\section{Introduction}

\begin{figure}[t!]
\centering
\includegraphics[width=0.95\linewidth]{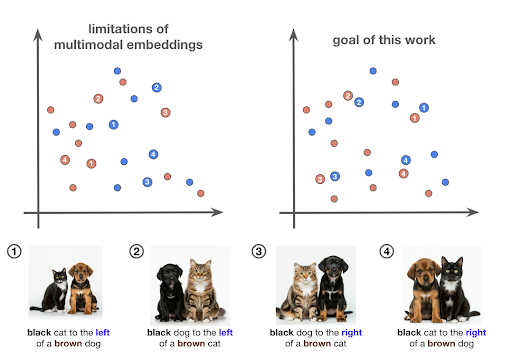}
\vspace{-0.2cm}
\caption{{\bf Summary of results.} Left: Previous state-of-the-art results in multimodal embeddings have limited understanding of the composition of images \cite{pmlr-v139-radford21a,jia2021scalingvisualvisionlanguagerepresentation}. Right: The goal of this work is to learn multimodal embeddings which reflect a strong understanding of the composition of visual and semantic information. Images and captions are red and blue, respectively.}
\label{fig:advertising-figure}
\end{figure}

Visual composition is a critical problem. Learning representations that capture how attributes and relationships of objects are represented is a critical task for any system that attempts to summarize the multitudes of concepts imbued within pixel space. 

Many SOTA multimodal models do not learn embeddings that have an understanding of the relationship of objects and descriptions \cite{OnoeDocci2024,yuksekgonul2023when}. One reason for this failure is that the visual representations are not rich enough to capture the interactions and relationships between various parts of an image or video.

Current approaches have focused on building new bespoke architectures that leverage multi-task learning through multiple losses or side data \cite{li2022blipbootstrappinglanguageimagepretraining,zeng2022multigrainedvisionlanguagepretraining}. Although these models achieve SOTA results on visual composition tasks \cite{yuksekgonul2023when}, it is unclear how scalable these approaches are given limitations on gathering high quality, specialized side information or the complexity of the architectures. 
    
Instead, we search for scalable solutions that focus on simplicity. The goal of this work is to employ a basic multimodal model solely employing a single contrastive learning objective \cite{pmlr-v139-radford21a,jia2021scalingvisualvisionlanguagerepresentation}, and attempt to improve the learned representations with minimal changes.
    
We posit that underlying visual architecture (e.g. ViT \cite{dosovitskiy2021imageworth16x16words}) contains sufficient parameters and scale to capture visual composition. Furthermore, we posit that contrastive learning is a sufficient training signal for multimodal learning so long as the target semantic embedding is rich enough. In particular, we ask if enriching the target semantic embedding is entirely sufficient to make a contrastive learning model capture visual composition.

We answer this question by identifying a small set of minimal changes to improve the target semantic embedding of a model. First, we recaption our training data using a strong multimodal foundation model (Figure \ref{fig:methods}. Second, we replace the text tower for CLIP \cite{pmlr-v139-radford21a,jia2021scalingvisualvisionlanguagerepresentation} with a powerful text-based foundation model. We find that these two changes are entirely sufficient to greatly improve the visual embedding representation. We measure this effect using standard benchmarks \cite{yuksekgonul2023when} and discuss more newer benchmarks \cite{OnoeDocci2024,saharia2022photorealistic} to showcase these performance gains. Notably, we find that for detailed image retrieval tasks, our approach improves the open-source CLIP model \cite{pmlr-v139-radford21a} recall @1 from 58.4\% to 94.5\%.

In summary, we focus on building simple and scalable techniques for improving compositionality in learned multimodal models focused on the problem of retrieval. We summarize our main contributions as follows:
\begin{itemize}[leftmargin=*]
    \item Rich visual captions may be automatically generated using foundational models with simple techniques to minimize hallucinations (Sec. \ref{section:methods}, \ref{section:docci}, and \ref{section:ablations}).
    \item Rich visual captions are sufficient for correcting many of the failures in compositional representation (Sec. \ref{section:aro}).
    \item COCO captioning benchmark \cite{coco} for measuring visual representations is saturated and subject to overfitting (Sec \ref{section:coco}). New benchmarks are necessary for measuring improvements \cite{OnoeDocci2024,yuksekgonul2023when} (Sec \ref{section:docci}).
\end{itemize}

%% file: sections/methods.tex
\section{Methods}
\label{section:methods}

\begin{figure*}[ht]
\centering
\includegraphics[width=0.9\linewidth]{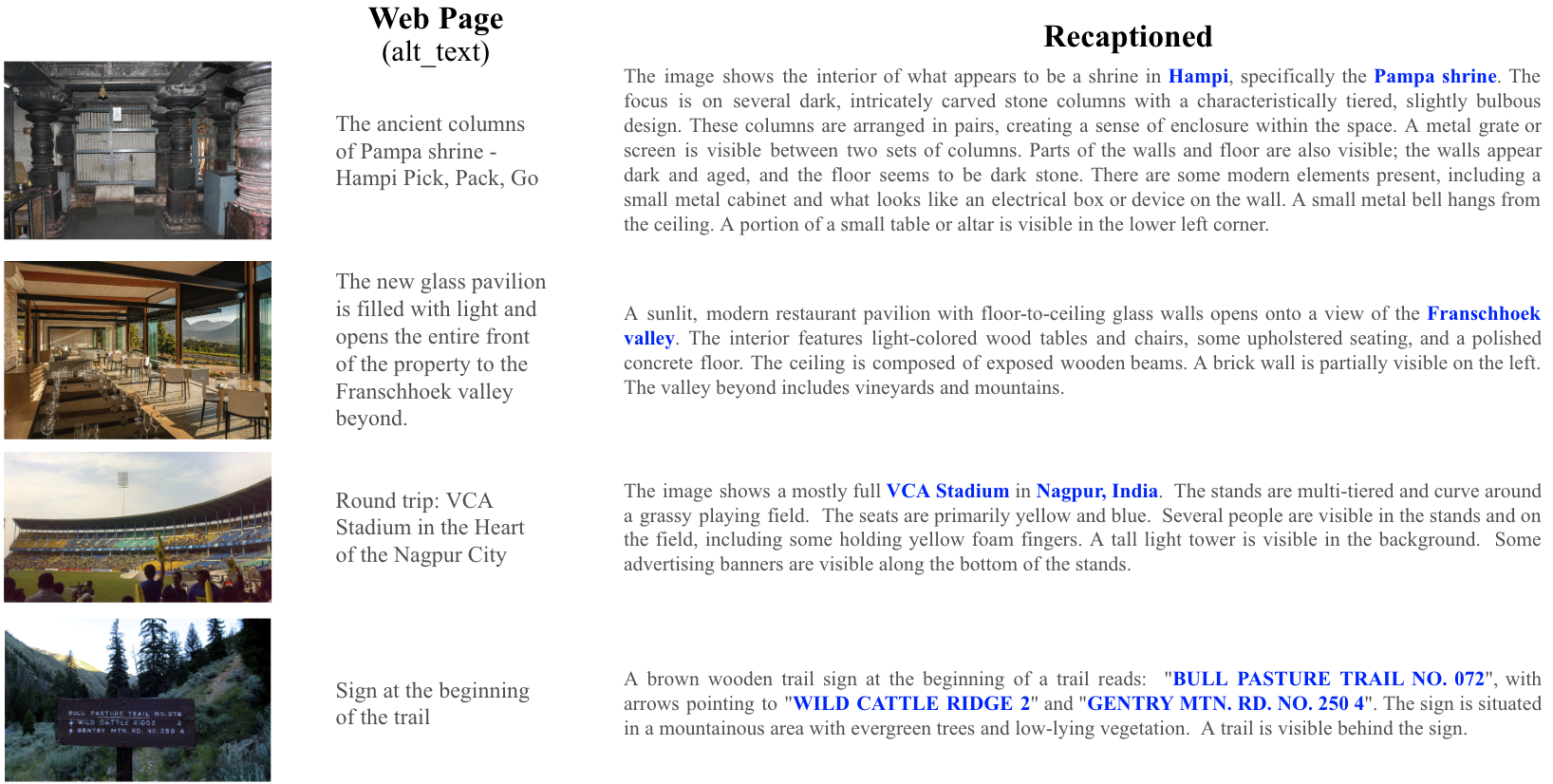}

\caption{{\bf Summary of methodology.} All images were recaptioned using a multimodal foundation model grounded on the image and the {\tt alt-text} for the image on the web page. In this case, we show captions generated using Gemini 1.5 Flash \cite{geminiteam2024gemini15unlockingmultimodal}. We highlight aspects of the new caption in {\bf \textcolor{blue}{blue}} that leverage the {\tt alt-text} or demonstrate a capability. Note how the generated captions leverage information provided by the {\tt alt-text} or perform OCR on the original image to improve the caption. Captions are for images from CC-12M \cite{changpinyo2021cc12m}.}
\label{fig:methods}
\end{figure*}

\begin{figure}[ht]
\centering
\includegraphics[width=0.98\linewidth]{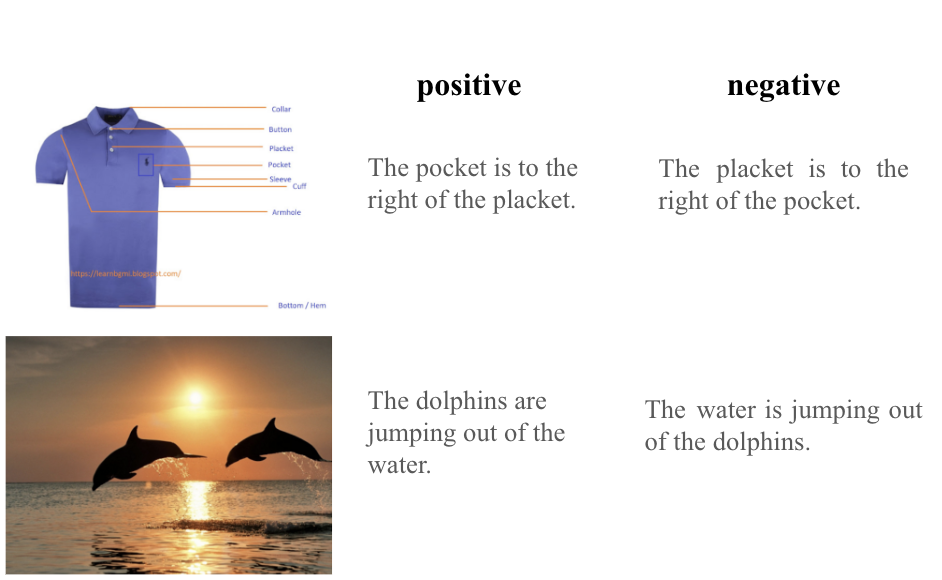}
\caption{{\bf Synthetic negative captions.} We prompt a foundation model \cite{geminiteam2024gemini15unlockingmultimodal} to generate 64 million synthetic positive and negative annotations. To generate the negative prompts, we provide few shot examples matching the style of ARO relations and attributes evaluation. Captions are for images from CC-12M \cite{changpinyo2021cc12m}.}
\label{fig:aro_negatives}
\end{figure}

Our method builds on the CLIP architecture proposed in \cite{pmlr-v139-radford21a,pmlr-v139-jia21b}. We introduce two key modifications:

\subsection{Semantic guidance with grounded recaptioning}
\label{section:method-recaptioning}

We start with an English-only subset of the WebLI dataset \cite{chen2022pali,dehghani2023scalingvisiontransformers22} consisting of 1B high quality images paired with freeform text scraped from the corresponding {\tt alt-text} within a web page.
We posit that the original {\tt alt-text} is noisy and a limiting factor in the performance of an multimodal model \cite{lai2024veclipimprovingcliptraining}.
Thus, we focus on creating a new set of captions to improve the {\tt alt-text}.

To enhance caption quality, we leverage Gemini 1.5 Flash \cite{reid2024gemini} to generate new captions. In early experiment, we provide the original image as well as the {\tt alt-text} and web page title, and prompt the model to generate a new caption that describes the underlying image. We iterated on the prompt (see Appendix for details) to arrive at a method that minimizes hallucinations while providing rich descriptions.

Figure \ref{fig:methods} summarizes our simple procedure. These examples showcase features in the recaptioning procedure. First, the grounding text supplied by the web page title and {\tt alt-text} provides rich grounding information which can be exploited by the model. Second, the multimodal model can employ OCR in the original image to improve the noisy {\tt alt-text} and correct for errors.

In early experiments, we additionally found significant gains by providing two forms of data augmentation. First, we performed sentence sampling, where by randomly selected subsets of caption sentences are used as targets for the model. Second we synthesized 2 million ``hard negative'' examples to our training mixture (Figure \ref{fig:aro_negatives}). We explored the choices of these two data augmentations in Section \ref{section:ablations}.

\begin{figure}
\centering
\includegraphics[width=0.85\linewidth]{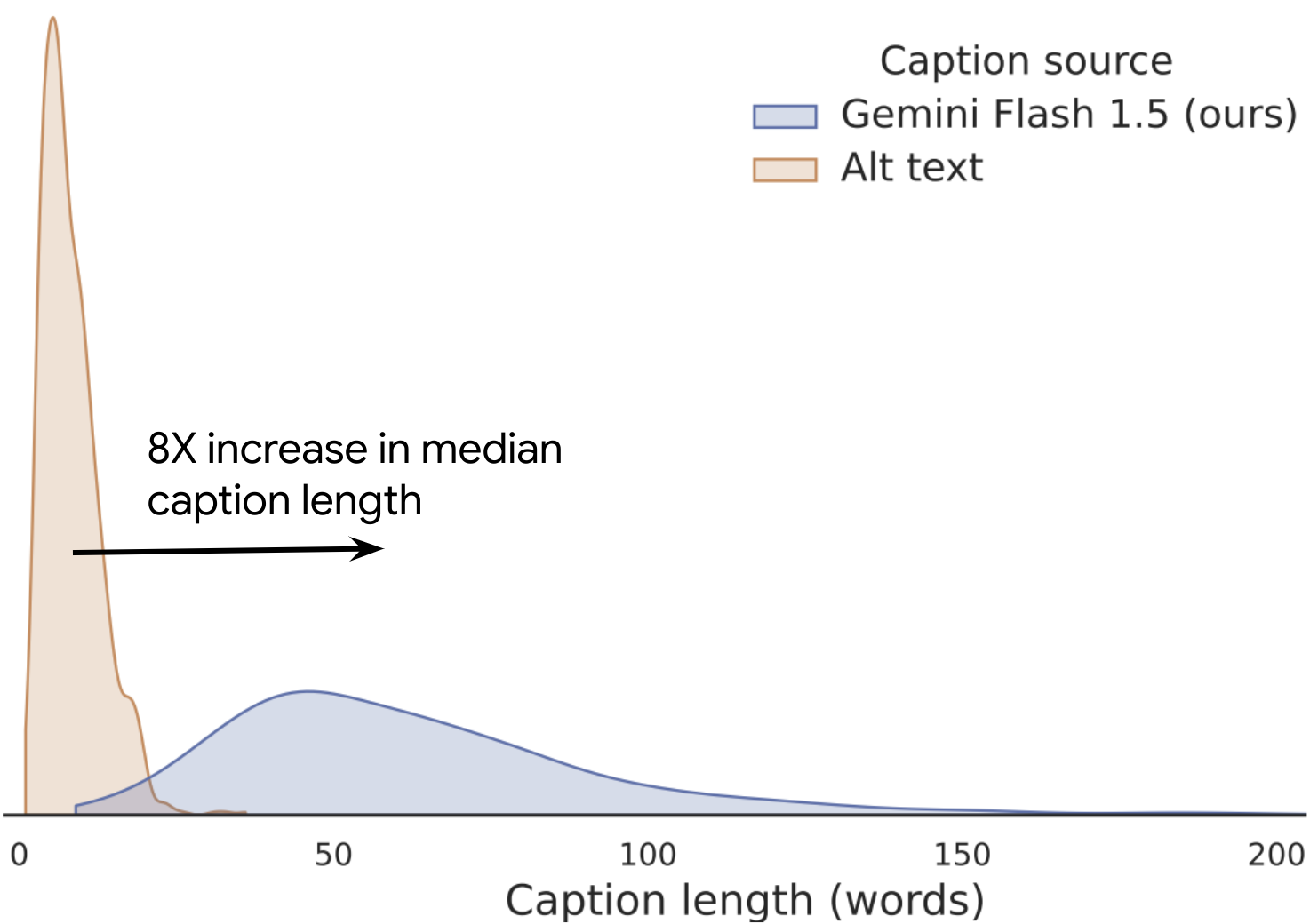}
\vspace{-0.2cm}
\caption{{\bf Recaptioning increases caption length by 8X.} While the median alt text caption length is just 7 words, the detailed captions generated by Gemini Flash 1.5 increase this to 57 words.}
\label{fig:caption-length}
\end{figure}

\begin{figure}
\centering
\includegraphics[width=0.95\linewidth]{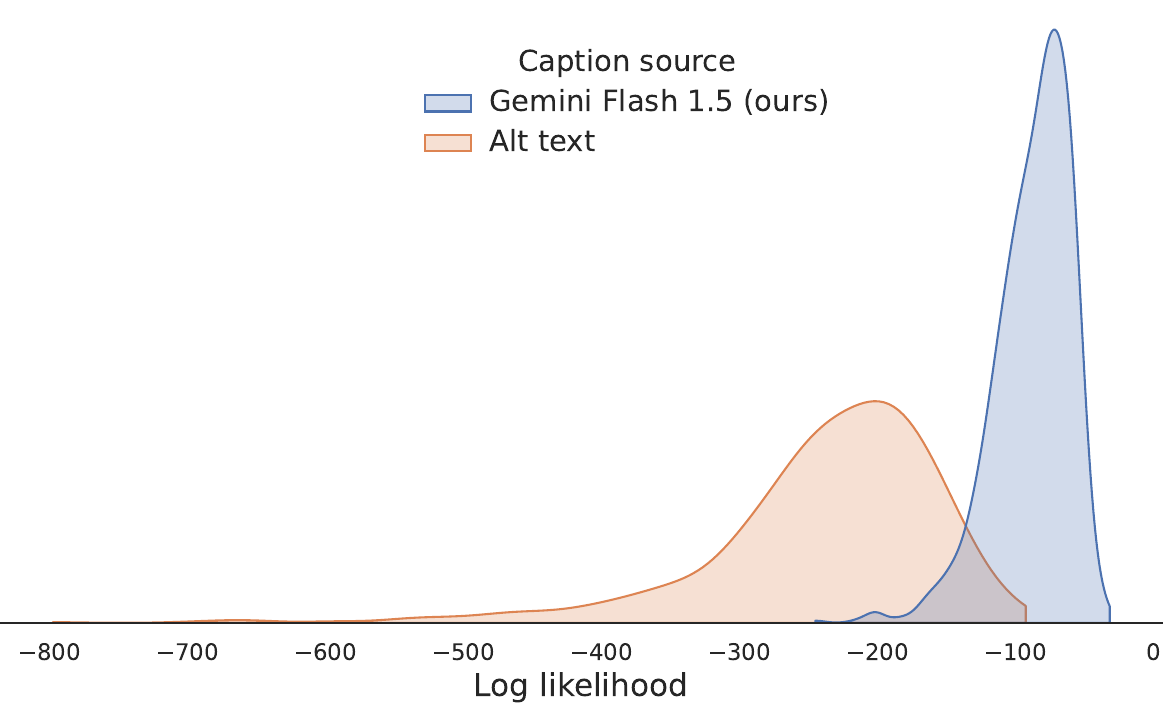}
\vspace{-0.2cm}
\caption{{\bf Recaptioning improves caption log-likelihood.} Alt text on the web is often unnatural (example: ``{\tt bigtimerush nyc 007}''), leading to low log-likelihood with a median of -223. In contrast, the captions from Gemini Flash 1.5 substantially improve median log-likelihood to -83, indicating that these captions are a lot closer to natural language and sentences than alt text.}
\label{fig:caption-loglikelihood}
\end{figure}

The resulting captions contain a mean of 57 words, increasing the effective caption length by about 8$\times$ of the {\tt alt-text} (Figure \ref{fig:caption-length}) and likewise significantly increasing the log-likelihood of the caption as measured by Gemini Pro 1.5 (Figure \ref{fig:caption-loglikelihood}).
In our ablations we experiment with an assortment of other captioning techniques. We generate a ``concise'' captions (26 $\pm 17$ words); we examine the importance of grounding the caption based on the {\tt alt-text} and page title; finally, we experiment with using a weaker LLM's as the captioning model (Section \ref{table:ablations} and see Appendix for additional details).

\subsection{Semantic guidance with a strong text encoder}

Instead of training a text encoder from scratch, we utilize the pretrained Gemini 1.5 Flash-8B \cite{geminiteam2024gemini15unlockingmultimodal} as the text encoder. We experiment with unfreezing various layers to allow for more flexibility in training the visual embedding. In early experiments, we explored unfreezing various layers of these pretrained text encoders and found that unfreezing last 4 layers of this model provides best overall performance relative to the additional compute cost during training. We repeated the above experiments with the open source Gemma2-2B  \cite{gemmateam2024gemma2improvingopen} as the text encoder, and likewise identified that unfreezing the last 4 layers provides the best trade off in terms of additional compute cost versus performance. This model is smaller than the Gemini model above, but yields comparable results.

%% file: sections/results.tex
\section{Results}
Our model is a standard two-tower CLIP encoder trained with a contrastive loss \cite{pmlr-v139-radford21a}. The image encoder is a
ViT-Base vision transformer with 86M parameters \cite{dosovitskiy2021imageworth16x16words}. Images are cropped
to 256$\times$256 pixels and represented as 256 patches of 16$\times$16 pixels. The encoders represent images and text
as $768$-dimensional embeddings.

For the text encoder, we experiment with two different configurations: In the first configuration, a bi-directional
text encoder with 12 layers and 297M parameters is trained from scratch matching
CoCa-Base \cite{yu2022}.
We explore also the use of a different text encoder based on a pre-trained Gemini 1.5 Flash-8B  \footnote{\url{https://ai.google.dev/gemini-api/docs/models/experimental-models}} and Gemma2-2B \cite{gemmateam2024gemma2improvingopen}. We keep most of the layers frozen,
and only fine-tune the last 4 layers. While the pre-trained text encoder
uses left-to-right context only, we switch to bi-directional context for the layers that are finetuned. 
In total, the model has 653M trainable parameters.
We perform 150,000 training steps with global batch size of 65,536 followed by fine-tuning for 500 training steps with reduced batch size of 4096 using data augmentation on the ``hard negative'' examples (Section \ref{section:method-recaptioning}).
The main model training is done on 256 accelerators and the fine-tuning is done with 16 accelerators. We use Adam optimizer \cite{Kingma2014AdamAM} with
a linear warm-up of the learning rate.

\subsection{Beyond a bag of words representation}
\label{section:aro}

\begin{table}[ht]
\centering
\begin{tabular}{lcccc}
\toprule
& relations & attributes \\
\midrule
Gemini-text \cite{geminiteam2024gemini15unlockingmultimodal} & 71\% & 82\% \\
Vera \cite{liu2023verageneralpurposeplausibilityestimation} & 62\% & 83\% \\
CLIP \cite{pmlr-v139-radford21a} & 59\% & 63\%  \\
CoCa \cite{yu2022} & 48\% & 50\% \\
NegCLIP \cite{yuksekgonul2023when} & 71\% & 81\% \\ 
BLIP $^\dagger$ \cite{li2022blipbootstrappinglanguageimagepretraining} & 59\% & 88\% \\
X-VLM $^\dagger$ \cite{zeng2022multigrainedvisionlanguagepretraining} & 73\% & 87\% \\
\midrule
Ours & 92\% & 94\% \\
\bottomrule
\end{tabular}
\vspace{0.15cm}
\caption{{\bf Our model improves upon bag of words.} Performance on Attribution, Relation, and Order (ARO) benchmark \cite{yuksekgonul2023when}. All numbers report classification macro accuracy. Chance rate is 50\%. Note that BLIP and X-VLM employ a second-stage binary classification in order to exhaustively identify the best candidate. Vera and Gemini-text only see the caption and not the images. \label{table:aro}}
\end{table}

\begin{table}[htbp]
    \centering
    \small 
    \setlength{\tabcolsep}{4pt} 
    \begin{tabular}{lccc|cc|cc}
        \toprule
        & \multicolumn{3}{c|}{\scriptsize replace} & \multicolumn{2}{c|}{\scriptsize swap} & \multicolumn{2}{c}{\scriptsize add} \\
        & \scriptsize O & \scriptsize A & \scriptsize R & \scriptsize O & \scriptsize A & \scriptsize O & \scriptsize A \\
        \midrule
        \scriptsize Human \cite{hsieh2023sugarcrepe}& \scriptsize 100\% & \scriptsize 99\% & \scriptsize 97\% & \scriptsize 99\% & \scriptsize 100\% & \scriptsize 99\% & \scriptsize 99\% \\
        \scriptsize Vera \cite{liu2023verageneralpurposeplausibilityestimation} & \scriptsize 49\% & \scriptsize 50\% & \scriptsize 49\% & \scriptsize 49\% & \scriptsize 49\% & \scriptsize 49\% & \scriptsize 50\% \\

        \scriptsize CLIP \cite{pmlr-v139-jia21b} & \scriptsize 94\% & \scriptsize 79\% & \scriptsize 65\% & \scriptsize 60\% & \scriptsize 62\% & \scriptsize 78\% & \scriptsize 72\% \\
        \scriptsize DC-XL \cite{gadre2023datacompsearchgenerationmultimodal} & \scriptsize 96\% & \scriptsize 85\% & \scriptsize 70\% & \scriptsize 65\% & \scriptsize 67\% & \scriptsize 91\% & \scriptsize 85\% \\

        \scriptsize LAION \cite{schuhmann2022laion5bopenlargescaledataset} & \scriptsize 97\% & \scriptsize 86\% & \scriptsize 72\% & \scriptsize 64\% & \scriptsize 72\% & \scriptsize 93\% & \scriptsize 86\% \\
        \scriptsize CapPa \cite{tschannen2023imagecaptionersscalablevision} & \scriptsize 92\% & \scriptsize 90\% & \scriptsize 87\% & \scriptsize 82\% & \scriptsize 88\% & \scriptsize 99\% & \scriptsize 99\% \\
        \scriptsize GPT-4V \cite{openai2024gpt4technicalreport} $^\dagger$ & \scriptsize 96\% & \scriptsize 94\% & \scriptsize 90\% & \scriptsize 83\% & \scriptsize 90\% & \scriptsize 92\% & \scriptsize 92\% \\
        \midrule
        \scriptsize Ours & \scriptsize 97\% & \scriptsize 94\% & \scriptsize 88\% & \scriptsize 89\% & \scriptsize 94\% & \scriptsize 95\% & \scriptsize 93\% \\
        \bottomrule
    \end{tabular}
    \caption{{\bf Results on SugarCrepe \cite{hsieh2023sugarcrepe}} O, A, and R stand for the object, attribute, and relation split, respectively. Chance rate is 50\%. Vera is a text-only model which does not see the images. $^\dagger$ GPT-4V sees both captions simultaneously and is not a two tower embedding model. LAION is {\tt xlm-roberta-large-ViT-H-14}. \label{table:sugarcrepe}}
\end{table}

A primary indication of the failure of multimodal representations can be observed in the relatively simple ARO \cite{yuksekgonul2023when} and SugarCrepe \cite{hsieh2023sugarcrepe} evaluation datasets. ARO artificially perturbs a set of captions from ``{\it the horse is eating grass}'' to ``{\it the grass is eating a horse}'' (i.e. relations) or ``{\it the paved road and the white house}'' to ``{\it the paved house and the white road}'' (i..e. attributes), and subsequently asks if the corresponding image is closer to the former or the latter caption. 
Chance rate is 50\% and a model that does respect the relations and attributes of an image -- such as a bag-of-words -- would correspondingly perform at chance rate. One of the most widely used multimodal baselines, CLIP \cite{pmlr-v139-radford21a}, achieves 59\% and 63\% accuracy on relations and attributes dimensions of ARO, indicating that the model is performing only slightly above a bag-of-words representation (Table \ref{table:aro}).

Admittedly, this task is artificial, and some captions can be correctly identified without examining an image (e.g., ``{\it grass cannot eat a horse}'') merely based on the language. 
To show this, we supplied Gemini 1.5 with two caption alternatives but \emph{no} image. This {\em blind} baseline achieves an accuracy of 71\% and 82\% on relation and attributes, respectively, outperforming CLIP. SugarCrepe improves upon this deficiency such that even text-only models specifically trained for verbal plausibility such as \cite{liu2023verageneralpurposeplausibilityestimation} only score at random chance levels. Apart from the more plausible negative captions, SugarCrepe is similar to ARO. We additionally show our results on SugarCrepe in Table  \ref{table:sugarcrepe}.

Several directions have been pursued for improving performance on these baselines (Table \ref{table:aro}, \ref{table:sugarcrepe}).
The ARO benchmark proposed a training augmentation method NegCLIP in which hard negatives are artificially added to the training set, which achieves 81\% and 71\% accuracy.
More importantly, several works have built bespoke architectures, derived from CLIP, which exploit localization information to better ground visual information.
Correspondingly, two different architectures achieves SOTA performance on relations (73\% \cite{zeng2022multigrainedvisionlanguagepretraining}) and attributes (88\% \cite{li2022blipbootstrappinglanguageimagepretraining}).
Both models achieve these results by tieing the image and text towers together, using a  grounded, cross-modal encoder \cite{zeng2022multigrainedvisionlanguagepretraining,li2022blipbootstrappinglanguageimagepretraining}. Although both methods employ a standard nearest neighbor lookup, they achieve SOTA results through a second-stage computation that exhaustively calculates a binary classification score for a set of candidate image-caption pairs. \cite{tschannen2023imagecaptionersscalablevision} replaces the contrastive objective with a captioning objective and shows this leads to gains on vision \& language tasks.

In comparison, our work makes no architectural changes and maintains a single contrastive training objective in a standard CLIP model. We do employ additional data augmentation with 64M synthetically generated ``hard negative'' examples (Section \ref{section:method-recaptioning}) and find that the addition of these data augmentations significantly improve compositional understanding (Section \ref{section:ablations}). Notably, our model achieves 92\% and 94\% accuracy on relations and attributes surpassing bespoke architectures that exhaustively search across examples. On SugarCrepe, we achieve state of the art or highly competitive results across results across all splits, even surpassing GPT-4V \cite{openai2024gpt4technicalreport} on all but one split. In the Appendix, we measure how our model uniformly outperforms competing methods across the assortment of fine-grained tasks comprising the ARO benchmark.

\subsection{Limitations of retrieval with COCO captions}
\label{section:coco}

\begin{figure*}[ht]
\centering
\begin{tabular}{cccc}
& {\tt \footnotesize{a couple of black cats}} & {\tt \footnotesize{several giraffes trotting and}} & {\tt \footnotesize{a cheese pizza on a plate}} \\ 
& {\tt \footnotesize{sitting on a window sill}} & {\tt \footnotesize{standing in a grassy fenced area}} & {\tt \footnotesize{on a table}} \\ 
\\ \vspace{0.02cm}
& \includegraphics[width=0.25\linewidth]{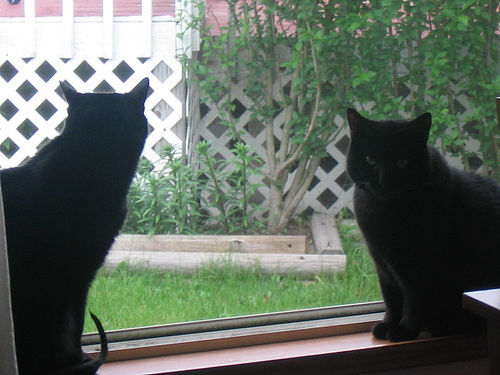} & \includegraphics[width=0.25\linewidth]{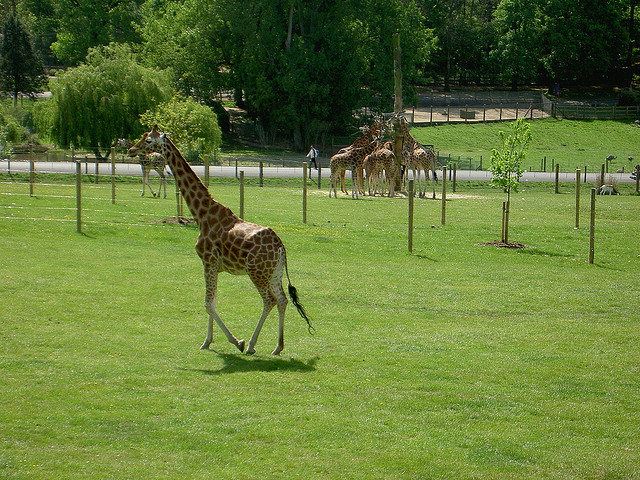} & \includegraphics[width=0.25\linewidth]{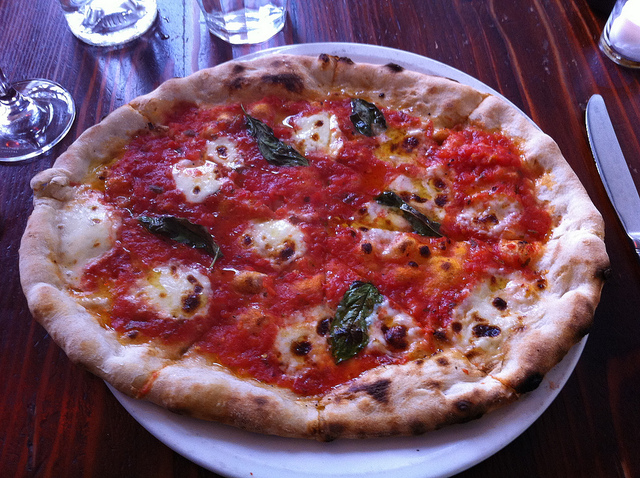} \\
& \includegraphics[width=0.25\linewidth]{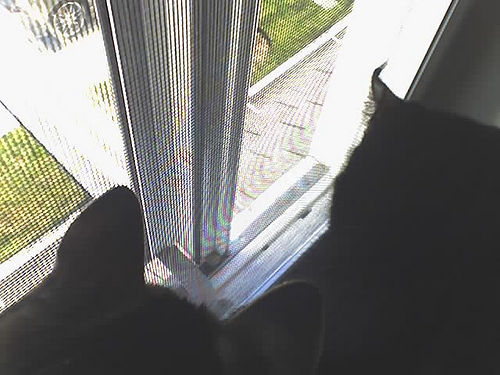} & \includegraphics[width=0.25\linewidth]{figures/coco-example-3-truth.png} & \includegraphics[width=0.25\linewidth]{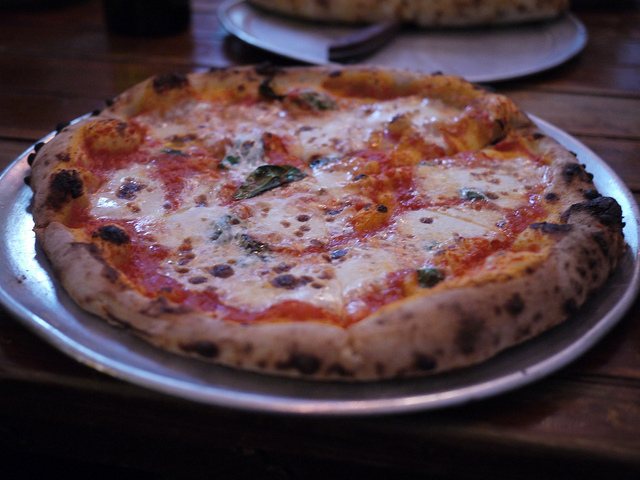} \\
\end{tabular}
\caption{{\bf Qualitative examples of limitations of image retrieval COCO captions.} Three examples showcasing where the model recalled image was incorrect, but is well described by the caption. {\bf Top and Middle Rows}: Caption and associated ground truth image. {\bf Bottom Row}: Top recalled image from our model. Note that the recalled images are well described by the caption corresponding to the original ground truth image.}
\label{fig:coco-examples}
\end{figure*}

Given the positive results on ARO, we asked whether these gains can be observed in image retrieval on the COCO dataset \cite{coco}. COCO contains detailed captions for 5,000 images and we ask how well nearest neighbor lookup in the embedding space performs for image retrieval. Our results compared to external baselines are in Table \ref{table:image-retrieval}.

As a strong baseline, we measure the performance of open source CLIP \cite{pmlr-v139-radford21a} and a CoCa \cite{yu2022} on image retrieval. We find a recall @1 of 37.8 and 47.5 on for CLIP and CoCa, respectively. Our model achieves the best reported recall of all models of similar size with a recall @ 1 of 56.3.

Although our result was stronger than prior numbers, we actually expected to see even more notable gains in our model given the high quality and scale of our data.
We investigated our model further by examining cases where our model is marked as retrieving the incorrect COCO image.
Figure \ref{fig:coco-examples} showcases three randomly selected examples. Note that although these examples are marked as incorrect, the image retrievals looks quite plausible. 
In fact, upon further inspection of the dataset, we observe that
many COCO captions contain similar images and captions, and in the context of image retrieval, we hypothesize that many image retrieval errors are  {\em incorrectly} scored as errors.

To test this hypothesis, we did a human annotation experiment to assess what fraction of images retrieved by our model are reasonable. We randomly select 500 failure cases and task 3 humans to annotate each of the failures to indicate whether the recalled image is an appropriate match to the prompt text. \footnote{During experimentation, we noticed that a notable fraction of the failure cases were because the ground truth caption for a given image was incorrect (i.e. in some examples, there is no image in the dataset which matches a given caption). We only assessed whether the recalled image matches the prompt caption, but this number might under report the true false failure rate due to some prompts having no appropriate image for our model to recall.} Using majority voting among the 3 human raters, we measured that 29.8\% of the image retrieval failures were appropriately marked failings of our model, but the other 70.2\% were acceptable matches to the query text. In other words, 70.2\% of all instances marked as failures were images that a majority of the human raters thought were appropriate image retrievals given the input caption.

These results indicate that a majority of the errors reported in Table \ref{table:image-retrieval} are due to noise and overly similar images or annotations in  COCO when employed as an image retrieval task. This result suggests that caution must be exercised when fine-tuning and evaluating on COCO, and motivates our interest to identify another evaluation benchmark that might provide a useful metric for measuring improvements in image retrieval that exercises a detailed semantic representation.

\subsection{Evaluation on detailed image caption retrieval}
\label{section:docci}

\begin{table}[t]
\centering
\begin{small}
\begin{tabular}{lcccc}
\toprule
& COCO & DOCCI-test & DOCCI-full \\ \midrule
CLIP \cite{pmlr-v139-radford21a}  & 37.8 & 58.4 & 46.2  \\
CoCa   \cite{yu2022} & 47.5 & 59.1 & 53.2 \\
BLIP \cite{li2022blipbootstrappinglanguageimagepretraining} & -- & 54.1 & -- \\ 
X-VLM \cite{zeng2022multigrainedvisionlanguagepretraining} & 56.1 & -- & -- \\
VeCLIP \cite{lai2024veclipimprovingcliptraining} & 48.9 & -- & -- \\
Long-CLIP \cite{zhang2024longclipunlockinglongtextcapability} & 40.4 & 45.2 & -- \\
MATE \cite{jang2024matemeetembedding} & -- & 73.4 & -- \\
TIPS \cite{anonymous2024tips} & 59.4 & 58.8 & -- \\  \midrule
Ours & 56.3 & 94.5 & 89.7 \\
\bottomrule
\end{tabular}
\end{small}
\caption{{\bf Image retrieval based on caption} on COCO \cite{coco} (5,000 images), DOCCI \cite{OnoeDocci2024} test split (5000 images) and the entire DOCCI dataset (15,847 images).
All numbers report recall @ 1. CoCa \cite{yu2022} is the base model. 
All models were {\em not} fine-tuned on COCO or DOCCI. 
DOCCI results for CLIP and CoCa reproduced from checkpoints. DOCCI results for BLIP and Long-CLIP reported in \cite{jang2024matemeetembedding}.
\label{table:image-retrieval}}
\end{table}

Given the positive results on the ARO benchmark but the inability to further discern gains on COCO, we investigated whether another retrieval benchmark may capture the gains in our approach. The DOCCI dataset was recently introduced in the context of image generation \cite{OnoeDocci2024} but has recently been explored for image retrieval. DOCCI contains 14,847 images with detailed {\it human-written} descriptions (see \cite{OnoeDocci2024} for details). Briefly, each caption contains on average 135.9 words across 7.1 sentences .
Importantly, DOCCI purposefully contains images and corresponding captions in which there exist slight changes and alterations to the pose, orientation, background, or action.
We construct an image retrieval task from this dataset and measure recall @ 1 for image retrieval. We report results on this dataset primarily in Table \ref{table:image-retrieval}, but see Tables \ref{table:ablations}, \ref{table:exploration}, \ref{table:sentsamp}, and \ref{table:hard-negatives} for detailed ablations.

Open-source CLIP baseline achieves 58.4\% recall. Notably, recently reported results on newer bespoke architectures with additional losses and side information achieve up to 73.4\% recall @ 1. The best performing model employs a novel projection module to better align text and visual representations \cite{jang2024matemeetembedding}. The authors additionally fine-tune on DOCCI train split at an increased image resolution of 448$\times$448 pixels to achieve their best result of 84.6\% recall @ 1.
We find that our model achieves 89.7\% recall @ 1 in a zero-shot manner using a standard CLIP training model but with our improved text supervision (but see Section \ref{section:ablations} for even higher numbers). We note that our model does not fine-tune on DOCCI train split, and uses a reduced resolution of 256$\times256$. We would expect that employing both well-known techniques, e.g. boosting image resolution and fine-tuning, would yield higher performance as well, but reserve such experiments for future work.

\subsection{Evaluation on zero-shot ImageNet classification}
\label{section:imagenet}

Prior work on embeddings and image retrieval have additionally evaluated on ImageNet and  variants \cite{deng2009imagenet}.
We emphasize that ImageNet emphasizes centrally-cropped, single objects within a scene, and {\em not} the composition of multiple ideas and objects.
Nonetheless, this dataset has proven to be a standard, reported metric in computer vision and, thus we apply our model to this task.
We find that our model achieves a zero-shot performance of 68.4\% top-1 accuracy (Table \ref{table:zero-shot}).
Although a reasonable zero-shot performance, we find nonetheless that this result is below other SOTA zero-shot ImageNet classification accuracies (Table \ref{table:zero-shot}).

One immediate discrepancy between our work and prior work is the source of training data. 
The open-source CLIP model employs a separate, proprietary training set that has never been revealed, so it is unclear how well this aligns with ImageNet classification task \footnote{We observe that CLIP achieves statistically significantly higher accuracy on a class-balanced version of the ImageNet training set than it does on the ImageNet validation set ($77.5\%$ train vs $76.2\%$ validation, two-proportion $z$-score$=6.2$, $p < 1 \times 10^{-9}$), indicating that some training data contamination maybe occurring for this model.}.
CoCa \cite{yu2022} (and other multimodal embedding models, e.g. LiT \cite{zhai2022litzeroshottransferlockedimage}, BASIC \cite{pham2023combinedscalingzeroshottransfer}) leverage a 50\% mixture of variations of the JFT dataset. 
In the case of JFT, the dataset comprises 3 billion images that have been annotated with a detailed class-hierarchy across 30K labels using a semi-automatic pipeline \cite{Zhai_2022_CVPR}.
We strongly suspect this fine-grained information contributes heavily to their overall performance.

To validate this hypothesis, we likewise introduce a 50\% mixture of JFT to our training set, and achieve 79.1\% zero-shot performance, suggesting that indeed the distribution of JFT is well tailored to ImageNet.
Although our result is still below the best reported zero-shot result of 82.6\%, we emphasize that there are a host of techniques that one could exploit to further boost this performance -- and were specifically employed by  \cite{yu2022,pmlr-v139-radford21a} \footnote{Two well-known techniques include fine-tuning models on high resolution data, and explicitly matching the captioning text used in ImageNet zero-shot evaluation in the training set. Both techniques provide additive gains for boosting performance \cite{yu2022}}.
We decided not to pursue this direction, and instead view the ImageNet results as demonstrating the model can indeed be performant for fine-grained discrimination if supplied the appropriate data. However, we view the topic of fine-grained classification as somewhat orthogonal to the central problem of building a model that better handles visual composition.

\begin{table}[t]
\small
\centering
\begin{tabular}{lc}
\toprule
& ImageNet ZS accuracy \\
\midrule
CLIP \cite{pmlr-v139-radford21a} & 76.2 \\
ALIGN \cite{pmlr-v139-jia21b} & 76.4 \\
CoCa \cite{yu2022} & 82.6 \\
\midrule
Ours & 68.4 \\
Ours (with JFT \cite{Zhai_2022_CVPR}) & 79.1 \\
\bottomrule
\end{tabular}
\caption{{\bf Zero-shot image classification results on ImageNet}. Prior work performance from \cite{yu2022}. CoCa report base model.\label{table:zero-shot}}
\end{table}

\subsection{Exploring the space of improved guidance}
\label{section:ablations}

\begin{table}[t]
\centering
\small
\begin{tabular}{cc|cc}
\toprule
Text Encoder & Text data & COCO & DOCCI-full \\
\midrule
Scratch   &  {\tt alt-text}  & 47.8 & 53.5  \\
Gemini-8B &  {\tt alt-text}  & 48.3 & 67.2 \\
Scratch   &  re-captioned & 46.5 & 75.6 \\
Gemini-8B &  re-captioned & 51.9 & 91.6 \\
Gemma-2B  &  re-captioned & 51.2 & 88.9 \\
\bottomrule
\end{tabular}
\vspace{0.15cm}
\caption{{\bf The importance of pretraining the text encoder and recaptioning.} We experimented with both Gemma \cite{gemmateam2024gemma2improvingopen} and Gemini \cite{geminiteam2024gemini15unlockingmultimodal} based pre-trained text encoders and see similar performance. Results were collected without hard-negative fine-tuning or random sentence sampling. \label{table:ablations}}
\end{table}

\begin{table}[]
    \centering
    \footnotesize
    \begin{tabular}{r|cc}
         & COCO & DOCCI-full \\
         \midrule
         Gemini 1.5 Flash \cite{geminiteam2024gemini15unlockingmultimodal} & 39.2 & 90.3 \\
         Gemini 1.5 Flash-8B \cite{geminiteam2024gemini15unlockingmultimodal} & 37.5 & 88.4 \\
    \end{tabular}\\
    \vspace{0.5cm}
    \begin{tabular}{cc|cc}
         length & grounded? & COCO & DOCCI-full \\
         \midrule
         default & \checkmark & 39.2 & 90.3 \\
         default & & 31.5 & 89.3 \\
         short & \checkmark & 39.0 & 85.2 \\
         {\it alt-text} & \checkmark & 38.4 & 65.7 \\
    \end{tabular}
\caption{{\bf Exploration of large-scale captioning} measured with recall @ 1 on image retrieval on COCO \cite{coco} and DOCCI \cite{OnoeDocci2024}.  All comparisons based on 100M captions from internal WebLI dataset \cite{chen2022pali}. {\bf Top}: Comparisons across different models for recaptioning. {\bf Bottom}: Comparison based on altering method for recaptioning. Grounded indicates whether {\tt alt-text} is supplied to the caption generation. Default and short caption length contain 133.4 and 354.8 words, respectively. Default captions used for rest of paper. Results were collected without hard-negative fine-tuning or random sentence sampling. \label{table:exploration}}
\end{table}

\begin{table}[]
    \centering
    \footnotesize 
    \begin{tabular}{c|cc}
        Sentence Sampling & COCO & DOCCI-full \\
         \midrule
          & 51.9 & 91.6 \\
         \checkmark & 56.3 & 93.0 \\
    \end{tabular}\\
\caption{{\bf Impact of random sentence sampling.} We randomly sample between one and ten sentences from our long generated captions. Doing so boosts recall on COCO and DOCCI. Results were collected without hard-negative fine-tuning. \label{table:sentsamp}}
\end{table}

\begin{table}[]
    \centering
    \footnotesize
    \begin{tabular}{c|cc|cc}
        Hard-neg. FT & COCO & DOCCI-full & ARO Attr. & ARO rel. \\
         \midrule
          & 51.9 & 91.6 & 82\% & 65\% \\
         \checkmark & 54.1 & 88.1 & 94\% & 93\% \\
    \end{tabular}\\
\caption{{\bf Impact of hard-negative fine-tuning.} We see that hard-negative fine-tuning significantly boosts results on ARO, slightly helps COCO, and slightly harms DOCCI. Results were collected without random sentence sampling. \label{table:hard-negatives}}
\end{table}

We examine the space of semantic guidance to better understand the additive benefits of each change in our model. Given that captioning 1B images is prohibitively expensive, we perform detailed ablations on a subset of 100M images for image retrieval on DOCCI and COCO (Tables \ref{table:ablations}, \ref{table:exploration}, \ref{table:sentsamp}, and \ref{table:hard-negatives}). 

Table \ref{table:ablations} indicates that training on recaptioning data alone notably boosts performance from 53.5 to 75.6 for recall@1 on DOCCI \cite{OnoeDocci2024}. We emphasize that the training images are identical; all that differs is the associated captions with the images. This small change results in 41\% relative performance gain. In parallel, we observe that using a rich text model as a target increases performance from 53.5 to 67.2, corresponding to 26\% relative performance boost. The two simple changes in tandem lead to additive gains on COCO and DOCCI.

Given the magnitude of benefits with recaptioning, we explored what features make for a good recaptioning. First, 
we examined how a less performant recaptioning models effects overall performance (Table \ref{table:exploration}, top). We observed that slightly 
drop in performance with a less performant recaptioning model.
Second, we examined how two key parameters in our recaptioning with Gemini 1.5 Flash effect overall quality. Namely, we removed grounding based on {\tt alt-text}, and we abbreviated length of the captions. We likewise observed that only slight degradations in overall performance on image retrieval (Table \ref{table:exploration}, bottom).
Finally, we asked how data augmentations for the captioning effected performance. In particular, we checked the impact from random sentence sampling (Table \ref{table:sentsamp}) and hard negatives fine-tuning (Table \ref{table:hard-negatives}). We observed that sentence sampling uniformly lead to better performance, especially on COCO. Conversely, removing the hard negatives did positively impact DOCCI at the significant harm to ARO and minor harm to COCO. We expect more work should be dedicated to finding the right balance of data augmentation in captioning pipelines in future work.

%% file: sections/related.tex
\section{Related Work}
\paragraph{Shortcomings of multimodal models.} Since the landmark publication of CLIP \cite{pmlr-v139-radford21a} as one of the first highly capable vision-language models, a number of important limitations have been observed in multimodal models. For instance, many image pairs are highly similar in CLIP feature space despite showing semantically meaningful differences (e.g., left-facing dog vs.\ right-facing dog). This leads to systematic failures in downstream models that use CLIP as an image encoder (although \cite{li2024erroneous} observed that better image-language alignment can sometimes alleviate this concern). Furthermore, vision-language models often behave like bag-of-word models \cite{yuksekgonul2023when}, as indicated through a surprising inability to understand compositions and relations (e.g., grass eating horse vs.\ horse eating grass). A number of approaches have been suggested for improving over current shortcomings; they are mostly focused on specialized architecture or data, and sometimes both.

\paragraph{Specialized architectures.} X-VLM \cite{zeng2022multigrainedvisionlanguagepretraining} proposes an architecture that incorporates bounding box prediction in order to improve the connection between visual input and text descriptions. The BLIP model \cite{li2022blipbootstrappinglanguageimagepretraining} is a popular specialized architecture trained jointly with three different objectives. CapPa \cite{tschannen2023imagecaptionersscalablevision} suggests image captioning as an alternative pretraining task leading to improved downstream performance. MATE \cite{jang2024matemeetembedding} identifies short captions as a problem (similar to our work); however instead of training models on longer captions (as we do) they propose an architecture that replaces the text encoder with a LLM-based one. Long-CLIP \cite{zhang2024longclipunlockinglongtextcapability} has a similar motivation and uses a bespoke architecture (modifying positional embedding, matching CLIP's primary component) in combination with one million long-text caption pairs. \cite{anonymous2024tips} is a concurrent work which uses multiple embedding heads, one for fine-grained tasks like ImageNet and the other for long form captions like DOCCI, in addition to a very large pretrained DINO g/14 image encoder and additional self-distillation and masking losses.

\paragraph{Data.} In contrast to the wealth of specialized architectures, the data side has not been explored as much---and frequently only in combination with a specialized architecture. Among the notable exceptions are \cite{patel2024tripletclip}, generating synthetic images from text-to-image models that are used as hard negatives in contrastive training. Furthermore, \cite{fan2023improvingcliptraininglanguage} use an LLM to rewrite (but not explicitly expand) existing captions as a form of data augmentation. In contrast to our work, the caption rewriting model is not visually grounded since it is a ``blind'' language model without image access. Finally, closest in spirit, \cite{lai2024veclipimprovingcliptraining} employed a multimodal model to more elaborate recaptioning pipeline but focused on boosting COCO retrieval.

%% file: sections/discussion.tex
\section{Discussion}

In this work we have highlighted the failures of multimodal embeddings. Current foundational models contain limited compositional information as observed by perturbations in the attributes and relationships as well as the failure to perform image retrieval on challenging captioning tasks  (Section \ref{section:aro}, \ref{section:coco}, and \ref{section:docci}).

This work addresses this problem by providing improved semantic guidance for multimodal models. In lieu of architectural innovations, this focuses on two simple and small changes that are highly scalable: (1) re-captioning training, and (2) improving the semantic target with a pretrained and powerful language model. These two improvements appear complimentary and result in a model that is able to perform 94.5 recall @1 on DOCCI image retrieval \cite{OnoeDocci2024} (Table \ref{table:image-retrieval}). 

More generally, this work highlights how bootstrapping data provides a scalable solution for constantly improving training data. Previous work had shown how simple procedures to bootstrap pseudo-labels may scale quite well and achieve state-of-the-art results on discrete classification tasks \cite{xie2020selftrainingnoisystudentimproves}. This work extends that endeavor by using pseudo-labels from other models to learn better embedding representations using contrastive learning.

As a simple application of this work, one benefit of improved alignment between visual and semantic representations is the opportunity to predict human preferences for image generation models.
For instance, in previous work Imagen \cite{saharia2022photorealistic} outperformed GLIDE \cite{nichol2022glidephotorealisticimagegeneration} in forced-choice human preference experiments for whether a caption is better aligned to the generated image.
As a pilot study, we asked whether our work may improve upon a strong baseline of the open-source CLIP model \cite{pmlr-v139-radford21a} for predicting these human preferences on the same dataset.
We observed that while open-source CLIP predicted the weighted human accuracy at 71.3\%, our model instead achieved 81.1\% weighted accuracy.
We also observed notable gains in terms of the ability of our model to count a small number of distinct objects.
We take these preliminary experiments as a suggestion for a potentially fruitful direction of exploration and reserve such a direction for future work.

This work has largely measured the quality of the model based on image retrieval in a shared embedding space. That said, a natural extension of this work is to investigate how these simple techniques may work in the context of an image captioning model (e.g. \cite{vinyals2016show}). Previous work has shown how image captioning as a training task may provide rich representation that may transfer to other visual tasks \cite{desai2021virtexlearningvisualrepresentations}. A natural question is to ask how such a training objective may additively improve multimodal representations \cite{yu2022}. In tandem, one could measure the quality of the presentation on captioning-based evaluations \cite{agrawal2016vqavisualquestionanswering} where challenges such as grounding and hallucination become more pronounced. We look forward to pursuing this direction in future work.

%% file: sections/appendix.tex
\appendix
\twocolumn[
\begin{center}
    {\bf {\Large Appendix}}
\end{center}
]

\section{Prompts used for recaptioning}
\label{appendix:recaptioning}
For gathering our main set of 1B image captions, we used the following prompt.

\begin{quote}
\itshape
\enquote{The image presented came from a web page called: {\tt page title} and had the alt-text: {\tt alt-text}. Please describe what is in the image using the alt-text and the page title as a guide to ground your response. For example, if the alt-text contains a specific brand name, use that brand name in your output. Please be descriptive but concise. DO NOT make things up. If you can't tell something with certainty in the image, simply don't say anything about it.}
\end{quote}

For gathering our 100M ablation set of very brief captions, we used the following prompt.

\begin{quote}
\itshape
\enquote{The image presented came from a web page called: {\tt page title} and had the alt-text: {\tt alt-text}. Please very briefly describe what is in the image using the alt-text and the page title as a guide to ground your response. For example, if the alt-text contains a specific brand name, use that brand name in your output. Please describe what is in the image but be extremely concise in your response. I want to emphasize how important it is to be VERY brief. DO NOT make things up. If you can't tell something with certainty in the image, simply don't say anything about it."}
\end{quote}

For gathering our 100M ablation set of captions without conditioning on alt-text or page title, we used the following prompt.

\begin{quote}
\itshape
\enquote{Please describe what is in the image. Please be descriptive but concise. DO NOT make things up. If you can't tell something with certainty in the image, simply don't say anything about it.}
\end{quote}

\section{Image to text retrieval results}

All prior results were given to as text $\rightarrow$ image recall. Following prior work, we also evaluated image $\rightarrow$ text recall.
\begin{table}[H]
\centering
\begin{small}
\begin{tabular}{lcccc}
\toprule
& COCO & DOCCI-test & DOCCI-full \\ \midrule
CLIP \cite{pmlr-v139-radford21a}  & 58.4 & 55.6 & 41.9 \\
CoCa   \cite{yu2022} & 63.8 & 56.7 & 51.8 \\
BLIP \cite{li2022blipbootstrappinglanguageimagepretraining} & -- & 54.7 & -- \\ 
X-VLM \cite{zeng2022multigrainedvisionlanguagepretraining} & 71.6 & -- & -- \\
VeCLIP \cite{lai2024veclipimprovingcliptraining} & 67.8 & -- & -- \\
Long-CLIP \cite{zhang2024longclipunlockinglongtextcapability} & 57.6 & 38.6 & -- \\
MATE \cite{jang2024matemeetembedding} & -- &  62.9 & -- \\ 
TIPS \cite{anonymous2024tips} & 74.0 & 57.2 & -- \\
\midrule
Ours & 76.2 & 95.9 & 91.3 &\\
\bottomrule
\end{tabular}
\end{small}
\caption{{\bf Caption retrieval based on image.} Same as Table \ref{table:image-retrieval} on COCO and DOCCI datasets, but on image $\rightarrow$ text retrieval. Note that COCO has multiple text labels for each image, making this task easier than text $\rightarrow$ image retrieval. \label{table:caption-retrieval}.}
\end{table}
\FloatBarrier

\section{Method details for caption statistics.}
\begin{figure}[]
\centering
\includegraphics[width=0.85\linewidth]{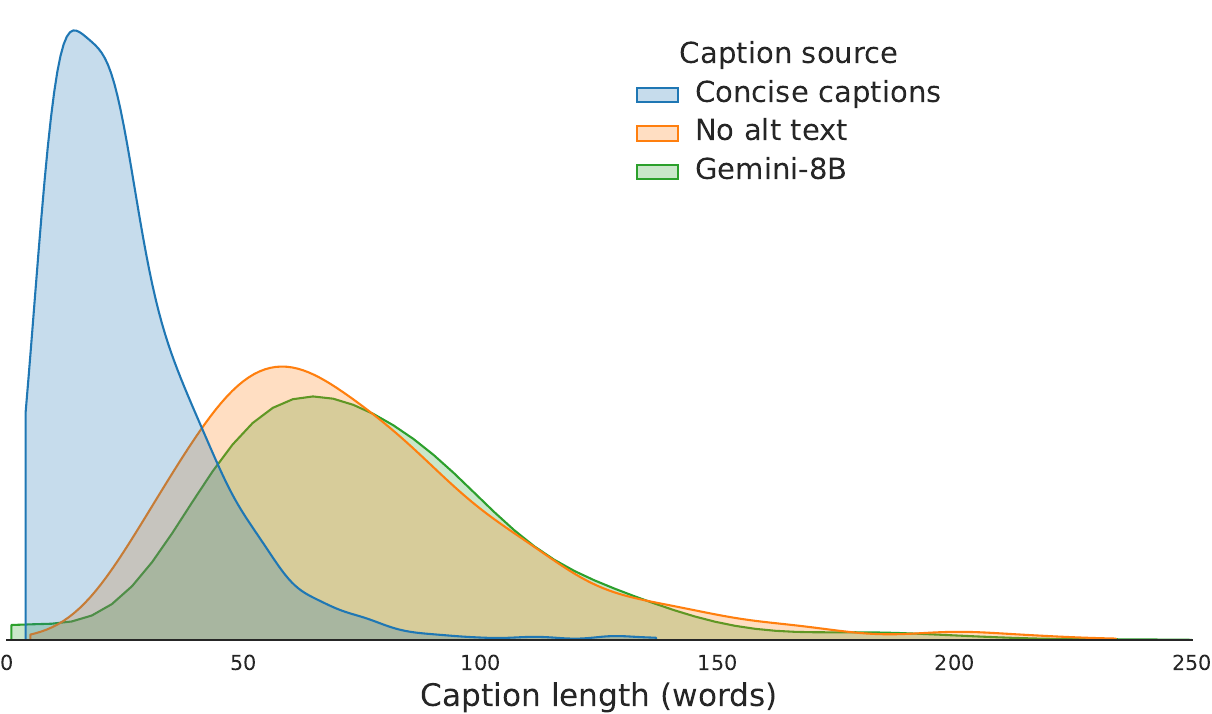}
\vspace{-0.2cm}
\caption{{\bf Caption length comparison for ablation captions.} Concise captions are significantly shorter than our default captions. Using Gemini-8B and removing alt-text conditioning have little impact on the length.}
\label{fig:caption-length-baselines}
\end{figure}

The plots from \cref{fig:caption-length,fig:caption-loglikelihood} are based on 1,000 random samples from each dataset (WebLI {\tt alt-text} vs.\ Gemini Flash 1.5 captions). The plots were generated via Seaborn's displot performing a kernel density estimate, setting cut=0 to avoid putting probability mass on negative caption lengths. The log-likelihood from \cref{fig:caption-loglikelihood} was obtained by scoring log-likelihood of {\tt alt-text} vs.\ Gemini Flash 1.5 captions on a random sample of 1,000 captions each. The model used for scoring was Gemini Pro 1.5 \citep{geminiteam2024gemini15unlockingmultimodal}, i.e., a larger and higher-quality model than the model used to generate captions. This conforms to the common practice in language modeling of using a large model to score text generation from a smaller model. The log-likelihood of a sequence of tokens $t = (t_1, ..., t_n)$ according to a language model parameterized by $\theta$ is calculated as follows:\\
$$\log p(t|\theta) = \displaystyle\sum_{i=1}^{n} \log p(t_i | t_1, t_2, ..., t_{i-1}; \theta)$$
For length statistics of ablation captions, see \cref{fig:caption-length-baselines}.

\section{Detailed results on ARO evaluation}

We report our performance across the fine-grained splits of ARO in Table \ref{appendix:table-relation-results}. We achieve high performance across most subsets of ARO. Notably, our performance on left/right is lower than other spatial relations, in particular much lower than similar relations such as above/below. We suspect there could be an issue with the ground truth labels for this subset. We had a human visualize and mark the correctness of a random sample of 100 left/right labels and determined that the ground truth for 43 out of 100 were either incorrect or ambiguous. Further investigation is warranted.
\input{sections/appendix-aro-table}

%% file: sections/appendix-aro-table.tex
\begin{table*}[tbh]
\centering
\small
\begin{tabular}{lrrrrrrr}
\toprule
{} &      CLIP &      NegCLIP &      CLIP-FT &      XVLM &      BLIP &      Flava &  Ours \\

\midrule
{\it Spatial Relationships} & 0.56 & 0.66 & 0.57 & 0.74 & 0.66 & 0.34 &  0.83\\\midrule
above           &  0.48 &  0.60 &  0.54 &  0.80 &  0.64 &  0.55 &   0.84\\
at              &  0.59 &  0.93 &  0.71 &  0.72 &  0.49 &  0.15 &    0.93\\
behind          &  0.56 &  0.29 &  0.34 &  0.82 &  0.77 &  0.28 &   0.80\\
below           &  0.56 &  0.46 &  0.48 &  0.74 &  0.69 &  0.44 &   0.78\\
beneath         &  0.80 &  0.70 &  0.70 &  0.80 &  0.70 &  0.40 &    0.80\\
in              &  0.63 &  0.89 &  0.63 &  0.73 &  0.72 &  0.09 &  0.99\\
in front of     &  0.54 &  0.75 &  0.70 &  0.66 &  0.55 &  0.78 &  0.85\\
inside          &  0.50 &  0.91 &  0.67 &  0.69 &  0.72 &  0.12 &    0.93\\
on              &  0.52 &  0.86 &  0.58 &  0.86 &  0.76 &  0.12 &  0.98\\
on top of       &  0.43 &  0.75 &  0.58 &  0.85 &  0.79 &  0.19 &   0.98\\
to the left of  &  0.49 &  0.50 &  0.50 &  0.52 &  0.51 &  0.50 & 0.59\\
to the right of &  0.49 &  0.50 &  0.50 &  0.52 &  0.49 &  0.51 &  0.61\\
under           &  0.64 &  0.43 &  0.54 &  0.86 &  0.73 &  0.27 &   0.69\\ \midrule
{\it Verbs} &   0.61 & 0.86 & 0.66 & 0.73 & 0.56 & 0.2 & 0.95 \\ \midrule
carrying          &  0.33 &  0.83 &  0.75 &  0.75 &  0.67 &  0.08 &    1.0\\
covered by        &  0.47 &  0.36 &  0.36 &  0.61 &  0.58 &  0.56 &    0.97\\
covered in        &  0.79 &  0.50 &  0.50 &  0.14 &  0.29 &  0.14 &    0.50\\
covered with      &  0.56 &  0.56 &  0.50 &  0.56 &  0.50 &  0.19 &    0.88\\
covering          &  0.39 &  0.58 &  0.45 &  0.67 &  0.55 &  0.06 &    0.97\\
cutting           &  0.75 &  0.83 &  0.83 &  0.67 &  0.25 &  0.00 &    1.0\\
eating            &  0.57 &  1.00 &  0.67 &  0.62 &  0.52 &  0.00 &    1.0\\
feeding           &  0.90 &  0.80 &  0.80 &  0.60 &  0.30 &  0.20 &    0.9\\
grazing on        &  0.10 &  0.90 &  0.30 &  0.60 &  0.40 &  0.50 &    1.0\\
hanging on        &  0.79 &  1.00 &  0.93 &  0.93 &  0.79 &  0.00 &    1.0\\
holding           &  0.58 &  0.97 &  0.79 &  0.67 &  0.44 &  0.27 &    1.0\\
leaning on        &  0.67 &  1.00 &  1.00 &  0.75 &  0.58 &  0.08 &    1.0\\
looking at        &  0.84 &  1.00 &  0.68 &  0.68 &  0.55 &  0.26 &    0.87\\
lying in          &  0.47 &  1.00 &  0.60 &  0.87 &  0.67 &  0.00 &    1.0\\
lying on          &  0.60 &  0.88 &  0.50 &  0.93 &  0.75 &  0.17 &    1.0\\
parked on         &  0.67 &  0.86 &  0.38 &  0.76 &  0.86 &  0.00 &    1.0\\
reflected in      &  0.64 &  0.71 &  0.57 &  0.50 &  0.43 &  0.43 &   0.86\\
resting on        &  0.38 &  0.85 &  0.23 &  0.92 &  0.54 &  0.15 &    1.0\\
riding            &  0.71 &  0.98 &  0.78 &  0.82 &  0.41 &  0.02 &    1.0\\
sitting at        &  0.62 &  1.00 &  0.88 &  0.88 &  0.46 &  0.00 &    1.0\\
sitting in        &  0.57 &  0.96 &  0.78 &  0.87 &  0.83 &  0.30 &    0.96\\
sitting on        &  0.58 &  0.97 &  0.78 &  0.94 &  0.73 &  0.14 &   0.99\\
sitting on top of &  0.50 &  0.90 &  0.80 &  1.00 &  0.80 &  0.10 &    1.0\\
standing by       &  0.67 &  0.92 &  0.67 &  0.83 &  0.67 &  0.67 &    0.92\\
standing in       &  0.73 &  0.98 &  0.69 &  0.69 &  0.49 &  0.05 &    1.0\\
standing on       &  0.60 &  1.00 &  0.63 &  0.83 &  0.73 &  0.06 &    1.0\\
surrounded by     &  0.64 &  0.71 &  0.64 &  0.71 &  0.64 &  0.79 &    0.93\\
using             &  0.84 &  1.00 &  1.00 &  0.68 &  0.58 &  0.00 &    1.0\\
walking in        &  0.70 &  1.00 &  0.70 &  0.60 &  0.50 &  0.00 &    1.0\\
walking on        &  0.79 &  1.00 &  0.79 &  0.84 &  0.42 &  0.05 &    1.0\\
watching          &  0.45 &  0.55 &  0.27 &  0.59 &  0.68 &  0.36 &    0.82\\
wearing           &  0.47 &  0.99 &  0.88 &  0.68 &  0.48 &  0.64 &   1.0\\
\midrule
\textbf{Overall} & 0.59 & 0.80 & 0.64 & 0.73 & 0.59 & 0.24 & 0.92\\

\bottomrule

\end{tabular}
\vspace{0.0cm}
\caption{{\bf Fine-grained results on ARO relations.}. Comparison results from \cite{yuksekgonul2023when}.}
\label{appendix:table-relation-results}

\end{table*}

%% file: paper.bbl
\begin{thebibliography}{40}
\providecommand{\natexlab}[1]{#1}
\providecommand{\url}[1]{\texttt{#1}}
\expandafter\ifx\csname urlstyle\endcsname\relax
  \providecommand{\doi}[1]{doi: #1}\else
  \providecommand{\doi}{doi: \begingroup \urlstyle{rm}\Url}\fi

\bibitem[Agrawal et~al.(2016)Agrawal, Lu, Antol, Mitchell, Zitnick, Batra, and
  Parikh]{agrawal2016vqavisualquestionanswering}
Aishwarya Agrawal, Jiasen Lu, Stanislaw Antol, Margaret Mitchell, C.~Lawrence
  Zitnick, Dhruv Batra, and Devi Parikh.
\newblock Vqa: Visual question answering, 2016.

\bibitem[Anonymous(2024)]{anonymous2024tips}
Anonymous.
\newblock {TIPS}: Text-image pretraining with spatial awareness.
\newblock In \emph{Submitted to The Thirteenth International Conference on
  Learning Representations}, 2024.
\newblock under review.

\bibitem[Changpinyo et~al.(2021)Changpinyo, Sharma, Ding, and
  Soricut]{changpinyo2021cc12m}
Soravit Changpinyo, Piyush Sharma, Nan Ding, and Radu Soricut.
\newblock {Conceptual 12M}: Pushing web-scale image-text pre-training to
  recognize long-tail visual concepts.
\newblock In \emph{CVPR}, 2021.

\bibitem[Chen et~al.(2022)Chen, Wang, Changpinyo, Piergiovanni, Padlewski,
  Salz, Goodman, Grycner, Mustafa, Beyer, et~al.]{chen2022pali}
Xi Chen, Xiao Wang, Soravit Changpinyo, AJ Piergiovanni, Piotr Padlewski,
  Daniel Salz, Sebastian Goodman, Adam Grycner, Basil Mustafa, Lucas Beyer,
  et~al.
\newblock Pali: A jointly-scaled multilingual language-image model.
\newblock \emph{arXiv preprint arXiv:2209.06794}, 2022.

\bibitem[Dehghani et~al.(2023)Dehghani, Djolonga, Mustafa, Padlewski, Heek,
  Gilmer, Steiner, Caron, Geirhos, Alabdulmohsin, Jenatton, Beyer, Tschannen,
  Arnab, Wang, Riquelme, Minderer, Puigcerver, Evci, Kumar, van Steenkiste,
  Elsayed, Mahendran, Yu, Oliver, Huot, Bastings, Collier, Gritsenko, Birodkar,
  Vasconcelos, Tay, Mensink, Kolesnikov, Pavetić, Tran, Kipf, Lučić, Zhai,
  Keysers, Harmsen, and Houlsby]{dehghani2023scalingvisiontransformers22}
Mostafa Dehghani, Josip Djolonga, Basil Mustafa, Piotr Padlewski, Jonathan
  Heek, Justin Gilmer, Andreas Steiner, Mathilde Caron, Robert Geirhos, Ibrahim
  Alabdulmohsin, Rodolphe Jenatton, Lucas Beyer, Michael Tschannen, Anurag
  Arnab, Xiao Wang, Carlos Riquelme, Matthias Minderer, Joan Puigcerver, Utku
  Evci, Manoj Kumar, Sjoerd van Steenkiste, Gamaleldin~F. Elsayed, Aravindh
  Mahendran, Fisher Yu, Avital Oliver, Fantine Huot, Jasmijn Bastings,
  Mark~Patrick Collier, Alexey Gritsenko, Vighnesh Birodkar, Cristina
  Vasconcelos, Yi Tay, Thomas Mensink, Alexander Kolesnikov, Filip Pavetić,
  Dustin Tran, Thomas Kipf, Mario Lučić, Xiaohua Zhai, Daniel Keysers,
  Jeremiah Harmsen, and Neil Houlsby.
\newblock Scaling vision transformers to 22 billion parameters, 2023.

\bibitem[Deng et~al.(2009)Deng, Dong, Socher, Li, Li, and
  Fei-Fei]{deng2009imagenet}
Jia Deng, Wei Dong, Richard Socher, Li-Jia Li, Kai Li, and Li Fei-Fei.
\newblock Imagenet: A large-scale hierarchical image database.
\newblock In \emph{2009 {IEEE} conference on computer vision and pattern
  recognition}, pages 248--255. {IEEE}, 2009.

\bibitem[Desai and Johnson(2021)]{desai2021virtexlearningvisualrepresentations}
Karan Desai and Justin Johnson.
\newblock Virtex: Learning visual representations from textual annotations,
  2021.

\bibitem[Dosovitskiy et~al.(2021)Dosovitskiy, Beyer, Kolesnikov, Weissenborn,
  Zhai, Unterthiner, Dehghani, Minderer, Heigold, Gelly, Uszkoreit, and
  Houlsby]{dosovitskiy2021imageworth16x16words}
Alexey Dosovitskiy, Lucas Beyer, Alexander Kolesnikov, Dirk Weissenborn,
  Xiaohua Zhai, Thomas Unterthiner, Mostafa Dehghani, Matthias Minderer, Georg
  Heigold, Sylvain Gelly, Jakob Uszkoreit, and Neil Houlsby.
\newblock An image is worth 16x16 words: Transformers for image recognition at
  scale, 2021.

\bibitem[Fan et~al.(2023)Fan, Krishnan, Isola, Katabi, and
  Tian]{fan2023improvingcliptraininglanguage}
Lijie Fan, Dilip Krishnan, Phillip Isola, Dina Katabi, and Yonglong Tian.
\newblock Improving clip training with language rewrites, 2023.

\bibitem[Gadre et~al.(2023)Gadre, Ilharco, Fang, Hayase, Smyrnis, Nguyen,
  Marten, Wortsman, Ghosh, Zhang, Orgad, Entezari, Daras, Pratt, Ramanujan,
  Bitton, Marathe, Mussmann, Vencu, Cherti, Krishna, Koh, Saukh, Ratner, Song,
  Hajishirzi, Farhadi, Beaumont, Oh, Dimakis, Jitsev, Carmon, Shankar, and
  Schmidt]{gadre2023datacompsearchgenerationmultimodal}
Samir~Yitzhak Gadre, Gabriel Ilharco, Alex Fang, Jonathan Hayase, Georgios
  Smyrnis, Thao Nguyen, Ryan Marten, Mitchell Wortsman, Dhruba Ghosh, Jieyu
  Zhang, Eyal Orgad, Rahim Entezari, Giannis Daras, Sarah Pratt, Vivek
  Ramanujan, Yonatan Bitton, Kalyani Marathe, Stephen Mussmann, Richard Vencu,
  Mehdi Cherti, Ranjay Krishna, Pang~Wei Koh, Olga Saukh, Alexander Ratner,
  Shuran Song, Hannaneh Hajishirzi, Ali Farhadi, Romain Beaumont, Sewoong Oh,
  Alex Dimakis, Jenia Jitsev, Yair Carmon, Vaishaal Shankar, and Ludwig
  Schmidt.
\newblock Datacomp: In search of the next generation of multimodal datasets,
  2023.

\bibitem[Hsieh et~al.(2023)Hsieh, Zhang, Ma, Kembhavi, and
  Krishna]{hsieh2023sugarcrepe}
Cheng-Yu Hsieh, Jieyu Zhang, Zixian Ma, Aniruddha Kembhavi, and Ranjay Krishna.
\newblock Sugarcrepe: Fixing hackable benchmarks for vision-language
  compositionality.
\newblock In \emph{Thirty-Seventh Conference on Neural Information Processing
  Systems Datasets and Benchmarks Track}, 2023.

\bibitem[Jang et~al.(2024)Jang, Kang, Lee, and Kim]{jang2024matemeetembedding}
Young~Kyun Jang, Junmo Kang, Yong~Jae Lee, and Donghyun Kim.
\newblock Mate: Meet at the embedding -- connecting images with long texts,
  2024.

\bibitem[Jia et~al.(2021{\natexlab{a}})Jia, Yang, Xia, Chen, Parekh, Pham, Le,
  Sung, Li, and Duerig]{pmlr-v139-jia21b}
Chao Jia, Yinfei Yang, Ye Xia, Yi-Ting Chen, Zarana Parekh, Hieu Pham, Quoc Le,
  Yun-Hsuan Sung, Zhen Li, and Tom Duerig.
\newblock Scaling up visual and vision-language representation learning with
  noisy text supervision.
\newblock In \emph{Proceedings of the 38th International Conference on Machine
  Learning}, pages 4904--4916. PMLR, 2021{\natexlab{a}}.

\bibitem[Jia et~al.(2021{\natexlab{b}})Jia, Yang, Xia, Chen, Parekh, Pham, Le,
  Sung, Li, and Duerig]{jia2021scalingvisualvisionlanguagerepresentation}
Chao Jia, Yinfei Yang, Ye Xia, Yi-Ting Chen, Zarana Parekh, Hieu Pham, Quoc~V.
  Le, Yunhsuan Sung, Zhen Li, and Tom Duerig.
\newblock Scaling up visual and vision-language representation learning with
  noisy text supervision, 2021{\natexlab{b}}.

\bibitem[Kingma and Ba(2014)]{Kingma2014AdamAM}
Diederik~P. Kingma and Jimmy Ba.
\newblock Adam: A method for stochastic optimization.
\newblock \emph{CoRR}, abs/1412.6980, 2014.

\bibitem[Lai et~al.(2024)Lai, Zhang, Zhang, Wu, Bai, Timofeev, Du, Gan, Shan,
  Chuah, Yang, and Cao]{lai2024veclipimprovingcliptraining}
Zhengfeng Lai, Haotian Zhang, Bowen Zhang, Wentao Wu, Haoping Bai, Aleksei
  Timofeev, Xianzhi Du, Zhe Gan, Jiulong Shan, Chen-Nee Chuah, Yinfei Yang, and
  Meng Cao.
\newblock Veclip: Improving clip training via visual-enriched captions, 2024.

\bibitem[Li et~al.(2022)Li, Li, Xiong, and
  Hoi]{li2022blipbootstrappinglanguageimagepretraining}
Junnan Li, Dongxu Li, Caiming Xiong, and Steven Hoi.
\newblock Blip: Bootstrapping language-image pre-training for unified
  vision-language understanding and generation, 2022.

\bibitem[Li et~al.(2024)Li, Koh, and Du]{li2024erroneous}
Siting Li, Pang~Wei Koh, and Simon~Shaolei Du.
\newblock On erroneous agreements of {CLIP} image embeddings.
\newblock \emph{arXiv preprint arXiv:2411.05195}, 2024.

\bibitem[Lin et~al.(2014)Lin, Maire, Belongie, Bourdev, Girshick, Hays, Perona,
  Ramanan, Doll{'{a} }r, and Zitnick]{coco}
Tsung{-}Yi Lin, Michael Maire, Serge~J. Belongie, Lubomir~D. Bourdev, Ross~B.
  Girshick, James Hays, Pietro Perona, Deva Ramanan, Piotr Doll{'{a} }r, and
  C.~Lawrence Zitnick.
\newblock Microsoft {COCO:} common objects in context.
\newblock \emph{CoRR}, abs/1405.0312, 2014.

\bibitem[Liu et~al.(2023)Liu, Wang, Wang, Smith, Choi, and
  Hajishirzi]{liu2023verageneralpurposeplausibilityestimation}
Jiacheng Liu, Wenya Wang, Dianzhuo Wang, Noah~A. Smith, Yejin Choi, and
  Hannaneh Hajishirzi.
\newblock Vera: A general-purpose plausibility estimation model for commonsense
  statements, 2023.

\bibitem[Nichol et~al.(2022)Nichol, Dhariwal, Ramesh, Shyam, Mishkin, McGrew,
  Sutskever, and Chen]{nichol2022glidephotorealisticimagegeneration}
Alex Nichol, Prafulla Dhariwal, Aditya Ramesh, Pranav Shyam, Pamela Mishkin,
  Bob McGrew, Ilya Sutskever, and Mark Chen.
\newblock Glide: Towards photorealistic image generation and editing with
  text-guided diffusion models, 2022.

\bibitem[Onoe et~al.(2024)Onoe, Rane, Berger, Bitton, Cho, Garg, Ku, Parekh,
  Pont-Tuset, Tanzer, Wang, and Baldridge]{OnoeDocci2024}
Yasumasa Onoe, Sunayana Rane, Zachary Berger, Yonatan Bitton, Jaemin Cho,
  Roopal Garg, Alexander Ku, Zarana Parekh, Jordi Pont-Tuset, Garrett Tanzer,
  Su Wang, and Jason Baldridge.
\newblock {DOCCI: Descriptions of Connected and Contrasting Images}.
\newblock In \emph{arXiv:2404.19753}, 2024.

\bibitem[OpenAI et~al.(2024)OpenAI, Achiam, Adler, Agarwal, Ahmad, Akkaya,
  Aleman, Almeida, Altenschmidt, Altman, Anadkat, Avila, Babuschkin, Balaji,
  Balcom, Baltescu, Bao, Bavarian, Belgum, Bello, Berdine, Bernadett-Shapiro,
  Berner, Bogdonoff, Boiko, Boyd, Brakman, Brockman, Brooks, Brundage, Button,
  Cai, Campbell, Cann, Carey, Carlson, Carmichael, Chan, Chang, Chantzis, Chen,
  Chen, Chen, Chen, Chen, Chess, Cho, Chu, Chung, Cummings, Currier, Dai,
  Decareaux, Degry, Deutsch, Deville, Dhar, Dohan, Dowling, Dunning, Ecoffet,
  Eleti, Eloundou, Farhi, Fedus, Felix, Fishman, Forte, Fulford, Gao, Georges,
  Gibson, Goel, Gogineni, Goh, Gontijo-Lopes, Gordon, Grafstein, Gray, Greene,
  Gross, Gu, Guo, Hallacy, Han, Harris, He, Heaton, Heidecke, Hesse, Hickey,
  Hickey, Hoeschele, Houghton, Hsu, Hu, Hu, Huizinga, Jain, Jain, Jang, Jiang,
  Jiang, Jin, Jin, Jomoto, Jonn, Jun, Kaftan, Łukasz Kaiser, Kamali,
  Kanitscheider, Keskar, Khan, Kilpatrick, Kim, Kim, Kim, Kirchner, Kiros,
  Knight, Kokotajlo, Łukasz Kondraciuk, Kondrich, Konstantinidis, Kosic,
  Krueger, Kuo, Lampe, Lan, Lee, Leike, Leung, Levy, Li, Lim, Lin, Lin, Litwin,
  Lopez, Lowe, Lue, Makanju, Malfacini, Manning, Markov, Markovski, Martin,
  Mayer, Mayne, McGrew, McKinney, McLeavey, McMillan, McNeil, Medina, Mehta,
  Menick, Metz, Mishchenko, Mishkin, Monaco, Morikawa, Mossing, Mu, Murati,
  Murk, Mély, Nair, Nakano, Nayak, Neelakantan, Ngo, Noh, Ouyang, O'Keefe,
  Pachocki, Paino, Palermo, Pantuliano, Parascandolo, Parish, Parparita,
  Passos, Pavlov, Peng, Perelman, de~Avila Belbute~Peres, Petrov,
  de~Oliveira~Pinto, Michael, Pokorny, Pokrass, Pong, Powell, Power, Power,
  Proehl, Puri, Radford, Rae, Ramesh, Raymond, Real, Rimbach, Ross, Rotsted,
  Roussez, Ryder, Saltarelli, Sanders, Santurkar, Sastry, Schmidt, Schnurr,
  Schulman, Selsam, Sheppard, Sherbakov, Shieh, Shoker, Shyam, Sidor, Sigler,
  Simens, Sitkin, Slama, Sohl, Sokolowsky, Song, Staudacher, Such, Summers,
  Sutskever, Tang, Tezak, Thompson, Tillet, Tootoonchian, Tseng, Tuggle,
  Turley, Tworek, Uribe, Vallone, Vijayvergiya, Voss, Wainwright, Wang, Wang,
  Wang, Ward, Wei, Weinmann, Welihinda, Welinder, Weng, Weng, Wiethoff,
  Willner, Winter, Wolrich, Wong, Workman, Wu, Wu, Wu, Xiao, Xu, Yoo, Yu, Yuan,
  Zaremba, Zellers, Zhang, Zhang, Zhao, Zheng, Zhuang, Zhuk, and
  Zoph]{openai2024gpt4technicalreport}
OpenAI, Josh Achiam, Steven Adler, Sandhini Agarwal, Lama Ahmad, Ilge Akkaya,
  Florencia~Leoni Aleman, Diogo Almeida, Janko Altenschmidt, Sam Altman,
  Shyamal Anadkat, Red Avila, Igor Babuschkin, Suchir Balaji, Valerie Balcom,
  Paul Baltescu, Haiming Bao, Mohammad Bavarian, Jeff Belgum, Irwan Bello, Jake
  Berdine, Gabriel Bernadett-Shapiro, Christopher Berner, Lenny Bogdonoff, Oleg
  Boiko, Madelaine Boyd, Anna-Luisa Brakman, Greg Brockman, Tim Brooks, Miles
  Brundage, Kevin Button, Trevor Cai, Rosie Campbell, Andrew Cann, Brittany
  Carey, Chelsea Carlson, Rory Carmichael, Brooke Chan, Che Chang, Fotis
  Chantzis, Derek Chen, Sully Chen, Ruby Chen, Jason Chen, Mark Chen, Ben
  Chess, Chester Cho, Casey Chu, Hyung~Won Chung, Dave Cummings, Jeremiah
  Currier, Yunxing Dai, Cory Decareaux, Thomas Degry, Noah Deutsch, Damien
  Deville, Arka Dhar, David Dohan, Steve Dowling, Sheila Dunning, Adrien
  Ecoffet, Atty Eleti, Tyna Eloundou, David Farhi, Liam Fedus, Niko Felix,
  Simón~Posada Fishman, Juston Forte, Isabella Fulford, Leo Gao, Elie Georges,
  Christian Gibson, Vik Goel, Tarun Gogineni, Gabriel Goh, Rapha Gontijo-Lopes,
  Jonathan Gordon, Morgan Grafstein, Scott Gray, Ryan Greene, Joshua Gross,
  Shixiang~Shane Gu, Yufei Guo, Chris Hallacy, Jesse Han, Jeff Harris, Yuchen
  He, Mike Heaton, Johannes Heidecke, Chris Hesse, Alan Hickey, Wade Hickey,
  Peter Hoeschele, Brandon Houghton, Kenny Hsu, Shengli Hu, Xin Hu, Joost
  Huizinga, Shantanu Jain, Shawn Jain, Joanne Jang, Angela Jiang, Roger Jiang,
  Haozhun Jin, Denny Jin, Shino Jomoto, Billie Jonn, Heewoo Jun, Tomer Kaftan,
  Łukasz Kaiser, Ali Kamali, Ingmar Kanitscheider, Nitish~Shirish Keskar,
  Tabarak Khan, Logan Kilpatrick, Jong~Wook Kim, Christina Kim, Yongjik Kim,
  Jan~Hendrik Kirchner, Jamie Kiros, Matt Knight, Daniel Kokotajlo, Łukasz
  Kondraciuk, Andrew Kondrich, Aris Konstantinidis, Kyle Kosic, Gretchen
  Krueger, Vishal Kuo, Michael Lampe, Ikai Lan, Teddy Lee, Jan Leike, Jade
  Leung, Daniel Levy, Chak~Ming Li, Rachel Lim, Molly Lin, Stephanie Lin,
  Mateusz Litwin, Theresa Lopez, Ryan Lowe, Patricia Lue, Anna Makanju, Kim
  Malfacini, Sam Manning, Todor Markov, Yaniv Markovski, Bianca Martin, Katie
  Mayer, Andrew Mayne, Bob McGrew, Scott~Mayer McKinney, Christine McLeavey,
  Paul McMillan, Jake McNeil, David Medina, Aalok Mehta, Jacob Menick, Luke
  Metz, Andrey Mishchenko, Pamela Mishkin, Vinnie Monaco, Evan Morikawa, Daniel
  Mossing, Tong Mu, Mira Murati, Oleg Murk, David Mély, Ashvin Nair, Reiichiro
  Nakano, Rajeev Nayak, Arvind Neelakantan, Richard Ngo, Hyeonwoo Noh, Long
  Ouyang, Cullen O'Keefe, Jakub Pachocki, Alex Paino, Joe Palermo, Ashley
  Pantuliano, Giambattista Parascandolo, Joel Parish, Emy Parparita, Alex
  Passos, Mikhail Pavlov, Andrew Peng, Adam Perelman, Filipe de Avila
  Belbute~Peres, Michael Petrov, Henrique~Ponde de Oliveira~Pinto, Michael,
  Pokorny, Michelle Pokrass, Vitchyr~H. Pong, Tolly Powell, Alethea Power,
  Boris Power, Elizabeth Proehl, Raul Puri, Alec Radford, Jack Rae, Aditya
  Ramesh, Cameron Raymond, Francis Real, Kendra Rimbach, Carl Ross, Bob
  Rotsted, Henri Roussez, Nick Ryder, Mario Saltarelli, Ted Sanders, Shibani
  Santurkar, Girish Sastry, Heather Schmidt, David Schnurr, John Schulman,
  Daniel Selsam, Kyla Sheppard, Toki Sherbakov, Jessica Shieh, Sarah Shoker,
  Pranav Shyam, Szymon Sidor, Eric Sigler, Maddie Simens, Jordan Sitkin,
  Katarina Slama, Ian Sohl, Benjamin Sokolowsky, Yang Song, Natalie Staudacher,
  Felipe~Petroski Such, Natalie Summers, Ilya Sutskever, Jie Tang, Nikolas
  Tezak, Madeleine~B. Thompson, Phil Tillet, Amin Tootoonchian, Elizabeth
  Tseng, Preston Tuggle, Nick Turley, Jerry Tworek, Juan Felipe~Cerón Uribe,
  Andrea Vallone, Arun Vijayvergiya, Chelsea Voss, Carroll Wainwright,
  Justin~Jay Wang, Alvin Wang, Ben Wang, Jonathan Ward, Jason Wei, CJ Weinmann,
  Akila Welihinda, Peter Welinder, Jiayi Weng, Lilian Weng, Matt Wiethoff, Dave
  Willner, Clemens Winter, Samuel Wolrich, Hannah Wong, Lauren Workman, Sherwin
  Wu, Jeff Wu, Michael Wu, Kai Xiao, Tao Xu, Sarah Yoo, Kevin Yu, Qiming Yuan,
  Wojciech Zaremba, Rowan Zellers, Chong Zhang, Marvin Zhang, Shengjia Zhao,
  Tianhao Zheng, Juntang Zhuang, William Zhuk, and Barret Zoph.
\newblock Gpt-4 technical report, 2024.

\bibitem[Patel et~al.(2024)Patel, Kusumba, Cheng, Kim, Gokhale, Baral, and
  Yang]{patel2024tripletclip}
Maitreya Patel, Abhiram Kusumba, Sheng Cheng, Changhoon Kim, Tejas Gokhale,
  Chitta Baral, and Yezhou Yang.
\newblock {TripletCLIP}: Improving compositional reasoning of {CLIP} via
  synthetic vision-language negatives.
\newblock \emph{arXiv preprint arXiv:2411.02545}, 2024.

\bibitem[Pham et~al.(2023)Pham, Dai, Ghiasi, Kawaguchi, Liu, Yu, Yu, Chen,
  Luong, Wu, Tan, and Le]{pham2023combinedscalingzeroshottransfer}
Hieu Pham, Zihang Dai, Golnaz Ghiasi, Kenji Kawaguchi, Hanxiao Liu, Adams~Wei
  Yu, Jiahui Yu, Yi-Ting Chen, Minh-Thang Luong, Yonghui Wu, Mingxing Tan, and
  Quoc~V. Le.
\newblock Combined scaling for zero-shot transfer learning, 2023.

\bibitem[Radford et~al.(2021)Radford, Kim, Hallacy, Ramesh, Goh, Agarwal,
  Sastry, Askell, Mishkin, Clark, Krueger, and Sutskever]{pmlr-v139-radford21a}
Alec Radford, Jong~Wook Kim, Chris Hallacy, Aditya Ramesh, Gabriel Goh,
  Sandhini Agarwal, Girish Sastry, Amanda Askell, Pamela Mishkin, Jack Clark,
  Gretchen Krueger, and Ilya Sutskever.
\newblock Learning transferable visual models from natural language
  supervision.
\newblock In \emph{Proceedings of the 38th International Conference on Machine
  Learning}, pages 8748--8763. PMLR, 2021.

\bibitem[Reid et~al.(2024)Reid, Savinov, Teplyashin, Lepikhin, Lillicrap,
  Alayrac, Soricut, Lazaridou, Firat, Schrittwieser, et~al.]{reid2024gemini}
Machel Reid, Nikolay Savinov, Denis Teplyashin, Dmitry Lepikhin, Timothy
  Lillicrap, Jean-baptiste Alayrac, Radu Soricut, Angeliki Lazaridou, Orhan
  Firat, Julian Schrittwieser, et~al.
\newblock Gemini 1.5: Unlocking multimodal understanding across millions of
  tokens of context.
\newblock \emph{arXiv preprint arXiv:2403.05530}, 2024.

\bibitem[Saharia et~al.(2022)Saharia, Chan, Saxena, Li, Whang, Denton,
  Ghasemipour, Gontijo~Lopes, Karagol~Ayan, Salimans,
  et~al.]{saharia2022photorealistic}
Chitwan Saharia, William Chan, Saurabh Saxena, Lala Li, Jay Whang, Emily~L
  Denton, Kamyar Ghasemipour, Raphael Gontijo~Lopes, Burcu Karagol~Ayan, Tim
  Salimans, et~al.
\newblock Photorealistic text-to-image diffusion models with deep language
  understanding.
\newblock \emph{Advances in neural information processing systems},
  35:\penalty0 36479--36494, 2022.

\bibitem[Schuhmann et~al.(2022)Schuhmann, Beaumont, Vencu, Gordon, Wightman,
  Cherti, Coombes, Katta, Mullis, Wortsman, Schramowski, Kundurthy, Crowson,
  Schmidt, Kaczmarczyk, and Jitsev]{schuhmann2022laion5bopenlargescaledataset}
Christoph Schuhmann, Romain Beaumont, Richard Vencu, Cade Gordon, Ross
  Wightman, Mehdi Cherti, Theo Coombes, Aarush Katta, Clayton Mullis, Mitchell
  Wortsman, Patrick Schramowski, Srivatsa Kundurthy, Katherine Crowson, Ludwig
  Schmidt, Robert Kaczmarczyk, and Jenia Jitsev.
\newblock Laion-5b: An open large-scale dataset for training next generation
  image-text models, 2022.

\bibitem[Team(2024{\natexlab{a}})]{geminiteam2024gemini15unlockingmultimodal}
Gemini Team.
\newblock Gemini 1.5: Unlocking multimodal understanding across millions of
  tokens of context, 2024{\natexlab{a}}.

\bibitem[Team(2024{\natexlab{b}})]{gemmateam2024gemma2improvingopen}
Gemma Team.
\newblock Gemma 2: Improving open language models at a practical size,
  2024{\natexlab{b}}.

\bibitem[Tschannen et~al.(2023)Tschannen, Kumar, Steiner, Zhai, Houlsby, and
  Beyer]{tschannen2023imagecaptionersscalablevision}
Michael Tschannen, Manoj Kumar, Andreas Steiner, Xiaohua Zhai, Neil Houlsby,
  and Lucas Beyer.
\newblock Image captioners are scalable vision learners too, 2023.

\bibitem[Vinyals et~al.(2016)Vinyals, Toshev, Bengio, and
  Erhan]{vinyals2016show}
Oriol Vinyals, Alexander Toshev, Samy Bengio, and Dumitru Erhan.
\newblock Show and tell: Lessons learned from the 2015 mscoco image captioning
  challenge.
\newblock \emph{IEEE transactions on pattern analysis and machine
  intelligence}, 39\penalty0 (4):\penalty0 652--663, 2016.

\bibitem[Xie et~al.(2020)Xie, Luong, Hovy, and
  Le]{xie2020selftrainingnoisystudentimproves}
Qizhe Xie, Minh-Thang Luong, Eduard Hovy, and Quoc~V. Le.
\newblock Self-training with noisy student improves imagenet classification,
  2020.

\bibitem[Yu et~al.(2022)Yu, Wang, Vasudevan, Yeung, Seyedhosseini, and
  Wu]{yu2022}
Jiahui Yu, Zirui Wang, Vijay Vasudevan, Legg Yeung, Mojtaba Seyedhosseini, and
  Yonghui Wu.
\newblock Co{C}a: Contrastive captioners are image-text foundation models.
\newblock \emph{Transactions on Machine Learning Research}, 2022.

\bibitem[Yuksekgonul et~al.(2023)Yuksekgonul, Bianchi, Kalluri, Jurafsky, and
  Zou]{yuksekgonul2023when}
Mert Yuksekgonul, Federico Bianchi, Pratyusha Kalluri, Dan Jurafsky, and James
  Zou.
\newblock When and why vision-language models behave like bags-of-words, and
  what to do about it?
\newblock In \emph{International Conference on Learning Representations}, 2023.

\bibitem[Zeng et~al.(2022)Zeng, Zhang, and
  Li]{zeng2022multigrainedvisionlanguagepretraining}
Yan Zeng, Xinsong Zhang, and Hang Li.
\newblock Multi-grained vision language pre-training: Aligning texts with
  visual concepts, 2022.

\bibitem[Zhai et~al.(2022{\natexlab{a}})Zhai, Kolesnikov, Houlsby, and
  Beyer]{Zhai_2022_CVPR}
Xiaohua Zhai, Alexander Kolesnikov, Neil Houlsby, and Lucas Beyer.
\newblock Scaling vision transformers.
\newblock In \emph{Proceedings of the IEEE/CVF Conference on Computer Vision
  and Pattern Recognition (CVPR)}, pages 12104--12113, 2022{\natexlab{a}}.

\bibitem[Zhai et~al.(2022{\natexlab{b}})Zhai, Wang, Mustafa, Steiner, Keysers,
  Kolesnikov, and Beyer]{zhai2022litzeroshottransferlockedimage}
Xiaohua Zhai, Xiao Wang, Basil Mustafa, Andreas Steiner, Daniel Keysers,
  Alexander Kolesnikov, and Lucas Beyer.
\newblock Lit: Zero-shot transfer with locked-image text tuning,
  2022{\natexlab{b}}.

\bibitem[Zhang et~al.(2024)Zhang, Zhang, Dong, Zang, and
  Wang]{zhang2024longclipunlockinglongtextcapability}
Beichen Zhang, Pan Zhang, Xiaoyi Dong, Yuhang Zang, and Jiaqi Wang.
\newblock Long-clip: Unlocking the long-text capability of clip, 2024.

\end{thebibliography}
